%% file: main.tex
\documentclass[fleqn,10pt]{wlscirep}
\usepackage[utf8]{inputenc}
\usepackage[T1]{fontenc}

\usepackage{caption}
\usepackage{cite}
\usepackage{amsmath,amssymb,amsfonts}
\usepackage{algorithm,algorithmic} 
\usepackage{booktabs}
\usepackage{graphicx}
\usepackage{textcomp}
\usepackage{float}
\usepackage{url} 
\usepackage{stfloats}

\title{DIPLI: Deep Image Prior Lucky Imaging for Blind Astronomical Image Restoration}

\author[1,*,$\dagger$]{Suraj Singh} 
\author[1,$\dagger$]{Anastasiia Batsheva}
\author[1,2]{Oleg Y. Rogov}
\author[3]{Ahmed Bouridane}

\affil[1]{Suraj Singh*, Anastasiia Batsheva and Oleg Y. Rogov are with the Skolkovo Institute of Science and Technology, Bolshoy blvd., 30/1, Moscow 121205, Russia}
\affil[2]{Oleg Y. Rogov is with AIRI, Moscow 121170, Russia}
\affil[3]{Ahmed Bouridane is with the College of Computing and Informatics, Computer Engineering Department, University of Sharjah, Sharjah 27272, UAE.}
\affil[$\dagger$]{Contributed equally}
\affil[*]{Corresponding author e-mail: Suraj.Singh@skoltech.ru}

\begin{abstract}
Modern image restoration and super-resolution methods utilize deep learning due to its superior performance compared to traditional algorithms. However, deep learning typically requires large labeled training datasets, which are rarely available in astrophotography. Deep Image Prior (DIP) bypasses this constraint by performing unsupervised optimization on a single image without training data; however, DIP often suffers from overfitting, artifact generation, and instability.
This work proposes DIPLI - a framework designed specifically for resolved, high-contrast astronomical targets that shifts from single-frame to multi-frame processing using the Back Projection technique, combined with dense optical flow estimation via the TVNet model, and replaces deterministic predictions with Monte Carlo estimation obtained through Stochastic Gradient Langevin Dynamics (SGLD). 
A comprehensive evaluation compares the method against the original DIP, the transformer-based model RVRT, and the diffusion-based model DiffIR2VR-Zero on synthetic data with ground truth, while comparing qualitatively against Lucky Imaging on real astronomical data. On synthetic datasets, DIPLI achieves the best perceptual fidelity scores (LPIPS in 12/12 and DISTS in 10/12 scenarios), while the diffusion-based DiffIR2VR-Zero achieves the best pixel-level distortion scores (PSNR in 9/12 and SSIM in 8/12 scenarios), consistent with the well-known perceptual--distortion trade-off in image restoration~\cite{Blau2018}. Compared to classical Lucky Imaging, the model requires far fewer input frames (7-13 versus thousands) and avoids the need for early stopping that limits standard DIP. Qualitative evaluation on real-world data of resolved solar-system objects, where ground truth is unavailable and domain shifts typically hinder generalization, suggests that the method appears to preserve fine detail while suppressing noise and artifacts.
\end{abstract}

\begin{document}

\flushbottom
\maketitle

\thispagestyle{empty}


\input{chapters/introduction}
\input{chapters/related_work}
\input{chapters/method}

\input{chapters/experiments}
\input{chapters/conclusion}

\section*{Code and Data Availability}
The source code for DIPLI, including all hyperparameter configurations used in this study, is available at \url{https://github.com/steelrail/DIPLI} (to be made public upon acceptance). Synthetic datasets are included in the repository. Real-world astronomical videos are available from the corresponding author on reasonable request.

\section*{Acknowledgment}
The corresponding author acknowledges the research support provided by Alex Gerko. The authors also express their gratitude to Andrew McCarthy for providing the astronomical photo dataset.

\section*{Author contributions statement}

Suraj Singh conceived the experiments, Suraj Singh and Anastasiia Batsheva conducted the experiments, Oleg Rogov and Ahmed Bouridane analysed the results. All authors reviewed the manuscript. 

\bibliography{sample}

\end{document}

%% file: chapters/introduction.tex
\section{Introduction}
\label{sec:introduction}

Imaging astronomical objects presents persistent challenges due to atmospheric turbulence, sensor noise, optical blur, and spatial distortions. These degradations stem from fundamental limitations in imaging hardware and observation conditions. While advanced equipment can mitigate some of these effects, software-based techniques, particularly super-resolution and image restoration, remain essential for reconstructing high-quality (HQ) images from low-quality (LQ) observations. Such methods are crucial not only for amateur astrophotographers and small research groups with limited resources but also for professional teams seeking to enhance data acquired under suboptimal conditions.

Among classical methods valued for their reliability and interpretability~\cite{Joshi_2010}, Lucky Imaging (LI)~\cite{Brandner_2016,Law_2006,WangLI2020} continues to be one of the most commonly used. LI reconstructs an HQ image from an unordered set of LQ frames by selecting a pivot frame, aligning all other frames to it via motion compensation, and subsequently averaging, as outlined in Algorithm \ref{a:liworkflow}. This method effectively suppresses noise and atmospheric distortions and remains actively used in telescope data pipelines due to its simplicity and reliability. However, the method also has notable limitations~\cite{Smith_2009}, including the need for a substantial number of source images, typically in the range of several thousand, to effectively suppress noise. In addition, its reliance on accurate pivot selection and motion estimation can hinder performance, as reported in~\cite{Oberkampf2006}. While recent advances in motion compensation and frame selection~\cite{Joshi2009} have improved its effectiveness, LI still struggles with highly dynamic scenes and faint celestial objects.

\begin{algorithm}[h]
\caption{Lucky Imaging Algorithm}
\label{a:liworkflow}
\begin{algorithmic}[1]
\REQUIRE $\{x_k\}_{k=1}^{K}$: LQ frames; $w(\cdot,u_k)$: spatial transform operator; quality metric $q(\cdot)$.
\ENSURE HQ estimate $x^*$.
\STATE $x_{\mathrm{pivot}} \leftarrow \arg\max_{k} q(x_k)$ 
\STATE $u_k \leftarrow \mathcal{M}\bigl(x_{\mathrm{pivot}}, x_k\bigr) \; \forall k$ - motion compensation 
\STATE $x^* = \frac{1}{K}\sum_{k=1}^{K} w(x_k, u_k)$
\RETURN $x^*$
\end{algorithmic}
\end{algorithm}

Deep Learning (DL) has demonstrated remarkable success in a variety of image restoration scenarios~\cite{Thomas_2021_CVPR,ExposureDL,XMMdenoiser,LocalizationWACV}. However, most DL models require extensive training data, which is typically unavailable in astrophotography. While pretrained models offer a potential solution, the significant domain gap between astronomical and natural images, on which most models are trained, makes them ineffective. Consequently, models capable of learning from limited data become necessary. One such approach is Deep Image Prior (DIP)~\cite{Ulyanov_2020}, which performs unsupervised optimization on a single image. Despite its appeal, DIP is highly sensitive, prone to overfitting, and lacks robustness in real-world conditions.

Several extensions of DIP have been proposed for multi-frame or video settings. Deep Video Prior~\cite{Lei2020} processes temporally ordered sequences by exploiting temporal consistency, while NAS-DIP~\cite{Chen_2020} and ISNAS-DIP~\cite{Arican_2021} search for task-specific architectures. These approaches are designed for temporally ordered sequences with smooth inter-frame transitions and have not been evaluated for the unordered, turbulence-degraded frame sets typical of astronomical observation. 

To address the limitations of existing methods, this work introduces \textbf{DIPLI}, a unified framework that combines the robustness of Lucky Imaging with the flexibility of Deep Image Prior. DIPLI operates on \emph{unordered} frame sets of resolved, high-contrast targets and fuses multi-frame information through a back-projection loss guided by estimated optical flows, without assuming temporal coherence.  The proposed model operates effectively with only 7–13 input frames by integrating Back Projection with optical flow estimated by the unsupervised TVNet model ~\cite{Fan2018}. To reduce overfitting and improve robustness, DIPLI replaces deterministic inference with Monte Carlo averaging using simplified Stochastic Gradient Langevin Dynamics (SGLD) ~\cite{welling2011bayesian} for sampling. This hybrid approach bridges classical and learning-based paradigms~\cite{DeepVarTPAMI}, achieving high-quality image reconstruction while minimizing data requirements and sensitivity to noise.

This work offers the following key contributions:
\begin{enumerate}
    \item The model extends DIP from single-frame to multi-frame processing using a back-projection loss with optical flow estimated by the unsupervised TVNet model, enabling reconstruction from as few as 7--13 frames.
    \item SGLD replaces the fragile early stopping heuristic of standard DIP with Monte Carlo averaging, eliminating the need for ground-truth-dependent stopping criteria.
    \item Comprehensive evaluation benchmarks the framework against the original DIP and state-of-the-art deep learning models (RVRT~\cite{liang2022recurrent} and DiffIR2VR-Zero~\cite{yeh2024diffir2vr}) on synthetic data using four complementary metrics (PSNR, SSIM~\cite{Wang2004}, LPIPS~\cite{zhang2018perceptual}, DISTS~\cite{DISTS}), and provides qualitative comparison on real astronomical data.
    \item Ablation studies examine the influence of the optical flow method, the number of input frames, and the SGLD noise strength on reconstruction quality.
\end{enumerate}

This paper is organized as follows: Section \ref{sec:related_work} reviews related work in astronomical imaging, classical restoration methods, and neural priors. Section \ref{sec:method} details the proposed DIPLI method. Section \ref{sec:experiments_discussion} presents experimental results, comparative evaluations, and ablation studies along with exhaustive analysis. Finally, section \ref{sec:conclusion} concludes with a summary and potential directions for future research.

%% file: chapters/related_work.tex
\section{Related Work}
\label{sec:related_work}

\paragraph{Inverse Imaging Problems.}
Reconstructing a HQ scene from a set of LQ observations which is formulated in Eq. ~\ref{eq:illposed} is generally an ill-posed Inverse Imaging Problem (IIP) ~\cite{Bertero_2020, Mccann_2017, Lucas_2018} since the degradation model (forward operator) is generally a non-bijective function, meaning that there could be more than one feasible solution for the HQ: 
\begin{equation}
\label{eq:illposed}
\text{LQ} = \text{Degradation Model}(\text{HQ}).
\end{equation}

While the form of the degradation model depends on the specific problem, a unified degradation framework is described in detail in Section~\ref{sec:method}. Some methods, such as Lucky Imaging, avoid directly using the inverse of a degradation model (backward operator). This approach ensures determinism and uniqueness of the result obtained. However, it does not consider the complete solution domain, and therefore does not guarantee global optimality. On the other hand, methods that seek to examine multiple candidates in the space of potential solutions $\text{P}$ face enormous computational challenges. In the case of the image domain, this space, or more accurately, distribution $P(\text{HQ}|\text{LQ})$ exhibits unpredictably complex statistics properties and is generally intractable. 

To limit the scope of the search, regularization mechanisms are typically implemented imposing constraints for a prior density of HQ~\cite{Gonzalez_2008, Buades_2005}. Traditional approaches rely on hand-crafted mathematical models~\cite{Chan_2005, Poonawala_2007, Qayyum_2021}, which often have limited discriminative capabilities.

Recently, DL models have emerged as a major advance in addressing ill-posed IIPs~\cite{Ongie_2020, Lucas_2018, Chen_2020}, surpassing the performance of methods based on hand-crafted priors. However, these DL-based approaches require large datasets of (LQ, HQ) pairs, with known ground truth images which can be challenging to acquire, particularly in astrophotography. Due to the lack of abundant datasets the field of astronomical image reconstruction still lies in an area of blind image reconstruction. 

\begin{figure}[t!]
   \centering
   \begin{minipage}{0.50\textwidth}
       \centering
       \includegraphics[width=\linewidth]{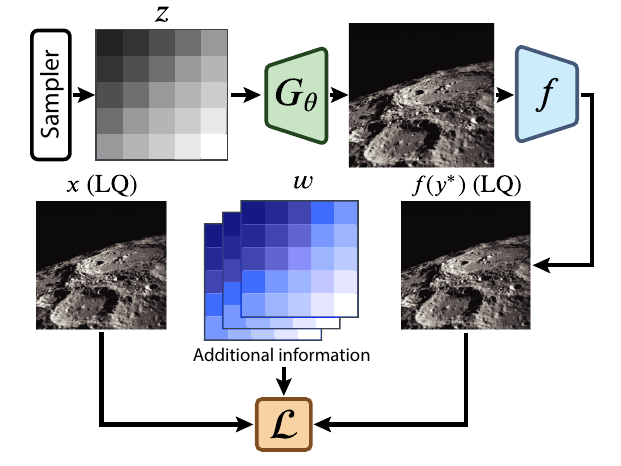}
       \caption{\textbf{Generalized UNNP reconstruction framework.} Before starting optimization, the \textit{sampler} generates a fixed input signal $z$ for the generator network $G_{\theta}$. $G_\theta$ learns to reconstruct the high-quality image $y^*$ based on the implicit regularization prior given by the network architecture. After a predefined forward degradation model $f$, the reconstruction $y^*$ is compared to a given set of LQ observations using the loss function $\mathcal{L}$ and additional information (such as optical flows, PSF, etc.) $\omega$.}
       \label{unnp_pipeline}
   \end{minipage}
   \hfill
   \begin{minipage}{0.43\textwidth}
   \centering
   \includegraphics[width=\linewidth]{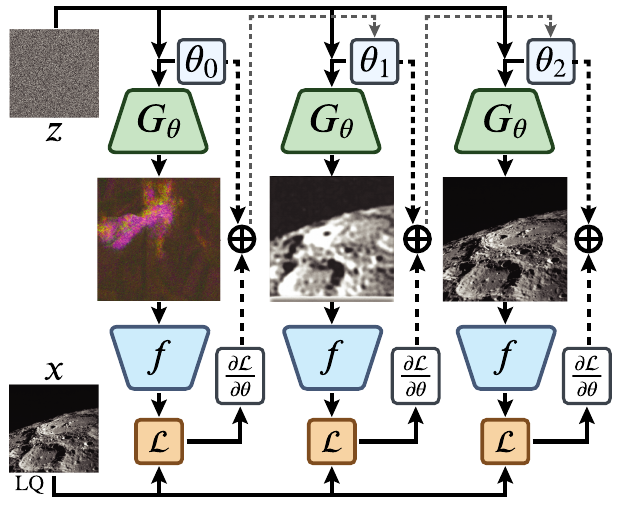}
   \caption{\textbf{Deep Image Prior optimization.} The goal of DIP optimization is to find parameters $\theta$ such that for a predetermined noise $z$, the output of the generator network $G_{\theta}$ will be a high-quality image $y^*$. The network $G_{\theta}$ is trained by minimizing the task-specific loss function $\mathcal{L}$ (usually mean squared error) between the given observation $x$ and the reconstruction $y^*$ distorted with the degradation process $f$.}
   \label{dip_optimization}
   \end{minipage}
\end{figure}

\paragraph{Deep Image Prior.}
The Untrained Neural Network Priors (UNNPs) framework ~\cite{Qayyum_2021, Rey_2022, Arican_2021, Ho_2020, Chen_2020, Uezato_2020, Liu_2021, Hong_2021} was originally proposed in~\cite{Ulyanov_2020}, bridging the gap between traditional hand-crafted priors and DL approaches. This generalized framework (Fig.\ref{unnp_pipeline}) can accurately estimate clean samples from a single corrupted measurement without prior knowledge of the ground truth, outperforming conventional hand-crafted optimization. UNNPs leverage the rich image statistics captured by randomly initialized convolutional neural networks (CNNs), with the network weights serving as a parameterization of the restored image.

The original framework for handling regularization tasks in inverse problems uses an untrained (randomly initialized parameters) CNN-based generator $G_\theta$ to solve various linear inverse problems, ranging from denoising to inpainting, super-resolution, and flash-no-flash reconstruction. The reconstruction $y^*$ is obtained by optimizing the parameters $\theta$ to maximize their likelihood given a task-specific observation model and a given degraded image (See Fig.\ref{dip_optimization} for reference):
\begin{equation}
y^* = \arg \min_y E(y, x),
\label{eq:DIP}
\end{equation}
where $E$ is a task-specific energy function. The motivation behind this approach is that the CNN architecture is biased towards natural images and can capture low-level image statistics without being explicitly trained on large-scale datasets. Instead, UNNPs can be applied directly to an observation.

In the case of $E(y, x) = \mathcal{L}_{\text{MSE}}(f(y), x)$, the optimization process can be denoted as:
\begin{equation}
\theta^* = \arg \min_\theta \mathcal{L}_{\text{MSE}}[f(G_\theta(z)), x],
\end{equation}
where $y^* = G_{\theta^*} (z)$ is the reconstruction of $y$, $f$ is a degradation function or an imaging system model, and $z \sim \mathcal{N}(0, I)$ is a fixed random input noise (latent code).

UNNPs exhibit high resistance to noise and strong ability to capture the true images. However, it has been shown~\cite{Ulyanov_2020} that as the number of iterations increases, the network also begins to memorize noise, which is known to be an early stopping challenge.

\paragraph{Bayesian inference.}
Consequently, selecting an appropriate coefficient for early stopping to mitigate this issue becomes necessary. Several heuristic techniques have been proposed to reduce this effect ~\cite{Heckel_2018, Dittmer_2019}. 

An alternative way to prevent overfitting is to apply Bayesian inference ~\cite{MacKay_1992, Radford_1995}, which involves integrating over the posterior distribution of HQ images (reparameterized via UNNP weights) with respect to the available LQ data:
\begin{align}
\label{e:lq-data}
\text{HQ}^* &= \text{degradation}^{-1}(\text{LQ}) \;\; \rightarrow \;\; \\
\text{HQ}^* &= \mathbb{E}_{p(\text{HQ}|\text{LQ})}
\left[\text{degradation}^{-1}(\text{LQ})\right].
\end{align}

To avoid the computation of this posterior average ~\cite{Cheng_2019}, in standard Markov Chain Monte Carlo (MCMC) ~\cite{Robert_2011} methods, the integral is replaced by a sample average of a Markov chain that converges to the true posterior average. However, convergence with MCMC techniques is significantly slower than Stochastic Gradient Descent (SGD)~\cite{SGD} for Deep Neural Networks (DNNs)\cite{li2024Convergence}. Therefore, it is more convenient to use SGLD~\cite{SGLD}, which transforms SGD into an MCMC sampler by injecting noise into gradient updates. Welling and Teh~\cite{welling2011bayesian} established the theoretical foundations of SGLD, showing that as the learning rate is annealed, the iterates converge to samples from the true posterior. Cheng et al.~\cite{Cheng_2019} further demonstrated that in the DIP setting, the posterior is well-behaved: the randomly initialized generator corresponds to a stationary Gaussian process prior, and SGLD sampling converges to the minimum mean square error (MMSE) solution without the need for early stopping. In practice, using a constant noise scale $\sigma_\xi$ equal to the learning rate - rather than the theoretically prescribed annealing schedule - yields approximate posterior samples that are effective for regularization, as demonstrated empirically in~\cite{Cheng_2019} and confirmed for the astronomical imaging domain in our experiments (Section~\ref{sec:experiments_discussion}, Fig.~\ref{sgld_plots}).

%% file: chapters/method.tex
\section{Method}
\label{sec:method}

In this section, we present DIPLI, an iterative approach to reconstructing a high-quality (HQ) scene \(y^*\) from a set of \(K\) distorted observations (low-quality or LQ frames) \(X = (x_1, x_2, \dots, x_K)\). Each LQ frame is modeled as:
\begin{equation}
x_k = f_k(y) + \eta_k, 
\quad \forall k \in \{1, \dots, K\},
\end{equation}
where \(\eta_k\) can be modeled as Gaussian noise $\eta_k \sim \mathcal{N}(0, \sigma_{\eta}^2 I)$ or consist of multiple components. To simulate realistic background light pollution in artificial datasets, a mixture of Gaussian and Poisson noise is employed.

\paragraph{Degradation model.}
The mapping \(f_k = d \circ h \circ \omega_k\) accounts for three major distortions:
\begin{itemize}
    \item \(d\): a downsampling operator, often Lanczos, chosen based on the desired output resolution.
    \item \(h\): a point spread function (PSF), which may be estimated during data acquisition, approximated by a Gaussian beam for sharper reconstructions, or omitted if unavailable.
    \item \(\omega_k\): a spatial distortion determined by the optical flow from a selected reference frame \(x_{\text{pivot}}\) to each \(x_k\).
\end{itemize}
To choose \(x_{\text{pivot}}\), the energy of the Laplacian~\cite{Marr1980} $q(x_k)$ is computed for every frame $x_k$, and the one maximizing this metric is selected:
\begin{equation}
x_{\text{pivot}} = \arg\max_k q(x_k).
\end{equation}
The optical flow~\cite{Horn_1981} between \(x_{\text{pivot}}\) and \(x_k\) is then estimated using TVNet~\cite{Fan2018}, which does not require pretraining and can be fine-tuned in an unsupervised manner to ensure smooth and coherent flow:
\begin{equation}
\omega_k = \mathcal{F}(x_{\text{pivot}}, x_k), 
\quad 
\Omega = (\omega_1, \omega_2, \dots, \omega_K).
\end{equation}

\paragraph{Multi-frame back-projection loss.}
Reconstruction of \(y\) from \(X\) is formulated as a single optimization problem that penalizes the discrepancy between the degraded reconstruction and each observed frame simultaneously, inspired by the back-projection principle~\cite{Hsieh_2011}:
\begin{gather}
y^* = \arg\min_{y \sim P(y)} \mathcal{L}(X, y; \Omega), \\
\mathcal{L}(X, y; \Omega) = \sum_{k=1}^K \bigl\| d \circ h \circ \omega_k (y) - x_k \bigr\|_2^2.
\end{gather}
In practice, optical flow estimates vary in reliability across the frame set. As a simple robustness measure, each frame's per-pixel contribution is weighted by a heuristic flow confidence map $c_k$. Following the classical observation that regions with rapidly varying flow fields are more likely to contain estimation errors~\cite{Horn_1981}, we use the spatial gradient magnitude of the flow as a proxy for local reliability: flow-field smoothness is a standard regularizer in variational methods~\cite{Sanchez2013}, and confidence measures based on local flow consistency are widely used in multi-frame fusion and stereo matching~\cite{Elad_1997}. For each flow field $\omega_k = (\omega_k^u, \omega_k^v)$, the confidence is computed as:
\begin{equation}
\label{eq:confidence}
c_k(p) = \exp\!\Bigl(-\alpha \bigl\|\nabla \omega_k(p)\bigr\|_F\Bigr),
\end{equation}
where $p$ denotes pixel position, $\nabla \omega_k(p)$ is the spatial Jacobian of the flow at $p$ computed via finite differences, $\|\cdot\|_F$ is the Frobenius norm, and $\alpha > 0$ is a scaling parameter. The weighted loss becomes:
\begin{equation}
\mathcal{L}_w(X, y; \Omega) = \sum_{k=1}^K \bigl\| c_k \odot \bigl(d \circ h \circ \omega_k(y) - x_k\bigr) \bigr\|_2^2,
\end{equation}
where $\odot$ denotes element-wise multiplication. At each optimization step, a random mini-batch of $K_b$ frames (with $K_b = \min(K, 4)$) is sampled from the full set to compute the loss gradient, which is a standard stochastic optimization practice that also has the practical effect of preventing any single frame's registration error from consistently dominating the gradient signal across iterations.
Rather than using a hand-crafted image prior \(P(y)\), an untrained neural network prior (UNNP)~\cite{Cheng_2019} is employed by letting
\begin{equation}
y = G_{\theta}(z), 
\quad z \sim \mathcal{N}(0, I).
\end{equation}
Thus, the problem becomes
\begin{equation}
y^* = G_{\theta^*}(z),
\quad
\theta^* = \arg\min_{\theta \sim P(\theta)} \mathcal{L}\bigl(X, G_{\theta}(z); \Omega\bigr).
\end{equation}

\paragraph{Variational inference.}
To mitigate overfitting, a variational inference framework is adopted to target the minimum mean square error (MMSE) solution~\cite{Cheng_2019}. Instead of returning a single optimal \(\theta^*\), the MMSE estimate is expressed as the expectation under the posterior distribution \(p(\theta \mid X)\):
\begin{equation}
y^*
= \mathbb{E}_{p(\theta \mid X)}\bigl[G_{\theta}(z)\bigr]
\approx
\frac{1}{N - n_0} \sum_{n=n_0+1}^N G_{\theta_n}(z),
\quad
\theta_n \sim p(\theta \mid X),
\end{equation}

Since direct sampling from \(p(\theta \mid X)\) is intractable, SGLD is employed:
\begin{gather}
\theta_0 \sim p_0(\theta), 
\\
\theta_n 
= \theta_{n-1} 
- \lambda_n \,\nabla_{\theta} \mathcal{L}\bigl(X, G_{\theta}(z);\Omega\bigr)\Bigr|_{\theta = \theta_{n-1}} 
+ \xi_n, 
\\
\xi_n \sim \mathcal{N}(0, \sigma_n^2 I).
\end{gather}
Cheng et al.~\cite{Cheng_2019} demonstrated that the randomly initialized DIP generator corresponds to a stationary Gaussian process prior, and that SGLD sampling in this setting converges to the MMSE solution. The multi-frame extension does not alter the SGLD mechanism: the back-projection loss aggregates multiple frames into a single loss function, so the parameter update retains the same SGLD form as in single-frame Bayesian DIP. Deviating from the theoretical annealing schedule $\sigma_n$, a constant \(\sigma_\xi\) equal to the learning rate is used to produce approximate posterior samples. After an initial warm-up phase of \(n_0\) iterations, the parameter vector \(\theta_n\) is treated as an approximate sample from \(p(\theta\mid X)\):
\begin{equation}
\theta_n \sim p(\theta \mid x_1, \dots, x_K),
\quad
n \ge n_0.
\end{equation}

In addition to SGLD noise on the parameters ($\xi_n$), a small independent perturbation is added to the latent code $z$ at each iteration:
\begin{equation}
G_{\theta_n}(z) \,\rightarrow\, G_{\theta_n}\bigl(z + z_n\bigr),
\quad
z_n \sim \mathcal{N}(0, \sigma_z^2 I).
\end{equation}
Both noise sources are always active simultaneously throughout optimization. The SGLD noise $\xi_n$ provides the primary regularization mechanism by enabling posterior exploration, while the latent perturbation $z_n$ provides a secondary smoothing effect. Cheng et al.~\cite{Cheng_2019} reported limited benefit from latent perturbation alone; we retain it following standard Bayesian DIP practice~\cite{Ulyanov2018DeepIP,CohenRandSmoothing} as it incurs negligible computational overhead.

\begin{algorithm}[H]
\caption{DIPLI Image Processing Workflow}
\label{a:workflow}
\begin{algorithmic}[1]
\REQUIRE LQ frames $\{x_k\}_{k=1}^{K}$; loss function $\mathcal{L}(\cdot, \cdot; \cdot)$, motion estimator $\mathcal{F}(\cdot)$; quality metric $q(\cdot)$, learning rate $\lambda_n$, SGLD strength $\sigma_{\xi}$; latent perturbations strength $\sigma_z$; total iterations $N$; warm-up iterations $n_0$; mini-batch size $K_b$.
\ENSURE HQ estimate $y^*$.
\STATE $x_{\mathrm{pivot}} \leftarrow \arg\max_{x_k} q(x_k)$
\STATE $\omega_k \leftarrow \mathcal{F}\bigl(x_{\mathrm{pivot}}, x_k\bigr) \; \forall k$
\STATE Compute confidence maps $c_k$ from $\omega_k$ via Eq.\,\eqref{eq:confidence} $\forall k$
\STATE $z\sim\mathcal{N}(0, I)$ \hfill \textit{(sampled once and reused)}
\STATE $\theta_0 \sim p_0(\theta)$; $\; y^* \leftarrow 0$
\FOR{$n \in \{1, \ldots, N\}$}
\STATE Sample mini-batch $\mathcal{B}_n \subset \{1,\ldots,K\}$, $|\mathcal{B}_n| = K_b$, uniformly at random
\STATE $z_n \sim \mathcal{N}(0,\sigma_z^2I)$
\STATE $\xi_n \sim \mathcal{N}(0, \sigma_{\xi}^2 I)$
\STATE $\theta_n \leftarrow \theta_{n-1}  - \lambda_n \,\nabla_{\theta} \mathcal{L}_w\bigl(\{x_k\}_{k\in\mathcal{B}_n}, G_{\theta}(z + z_n);\{\omega_k\}_{k\in\mathcal{B}_n}\bigr)\Bigr|_{\theta = \theta_{n-1}}  + \xi_n$
\IF{$n > n_0$}
\STATE $y^* \leftarrow y^* + G_{\theta_n}(z + z_n)$
\ENDIF
\ENDFOR
\RETURN $\frac{1}{N - n_0}y^*$ 
\end{algorithmic}
\end{algorithm}

%% file: chapters/experiments.tex
\section{Experiments and Discussion}
\label{sec:experiments_discussion}

\subsection{Dataset description}

The quality of the proposed DIPLI model was evaluated using two categories of experiments: one based on artificially generated data and the other on real-world data.

The real-world data consists of 15 videos capturing celestial objects, including the Moon, Mars, Saturn, Jupiter, and the Sun, provided at a resolution of $256 \times 256$ pixels. These videos are sourced from private collection and publicly available data. A primary challenge with real-world data is the absence of ground truth images for direct comparison, which restricts validation methods to no-reference quality metrics and human visual inspection. The energy of Laplacian~\cite{pang2017graph} was used for blind quality assessment, as it is a standard metric used in Lucky Imaging algorithms during data registration and image ranking processes.

The artificial dataset comprises artificially generated video sequences depicting the planetary images and data from Mars Exploration Rovers~\cite{ROVERmission} both at $256 \times 256$ pixels, alongside corresponding ground truth images at a resolution of $1024 \times 1024$ pixels. The presence of the ground truth images enables to evaluate the model's performance in terms of more trustworthy and interpretable reference-based metrics such as Peak Signal-to-Noise Ratio (PSNR), Structural Similarity Index (SSIM)~\cite{Wang2004}, along with Neural Network-based metrics namely LPIPS~\cite{zhang2018perceptual}, and DISTS~\cite{DISTS}. It should be noted that the ground-truth images are not used for network training, as the objective is to develop a blind super-resolution algorithm.

The artificial dataset was generated using the degradation model described in Section~\ref{sec:method}. Specifically, a high-resolution ground-truth image was progressively corrupted with noise and various spatial distortions to simulate the realistic process of video acquisition of a celestial object.
\begin{figure}[t!]
\centering
\includegraphics[width=0.9\linewidth]{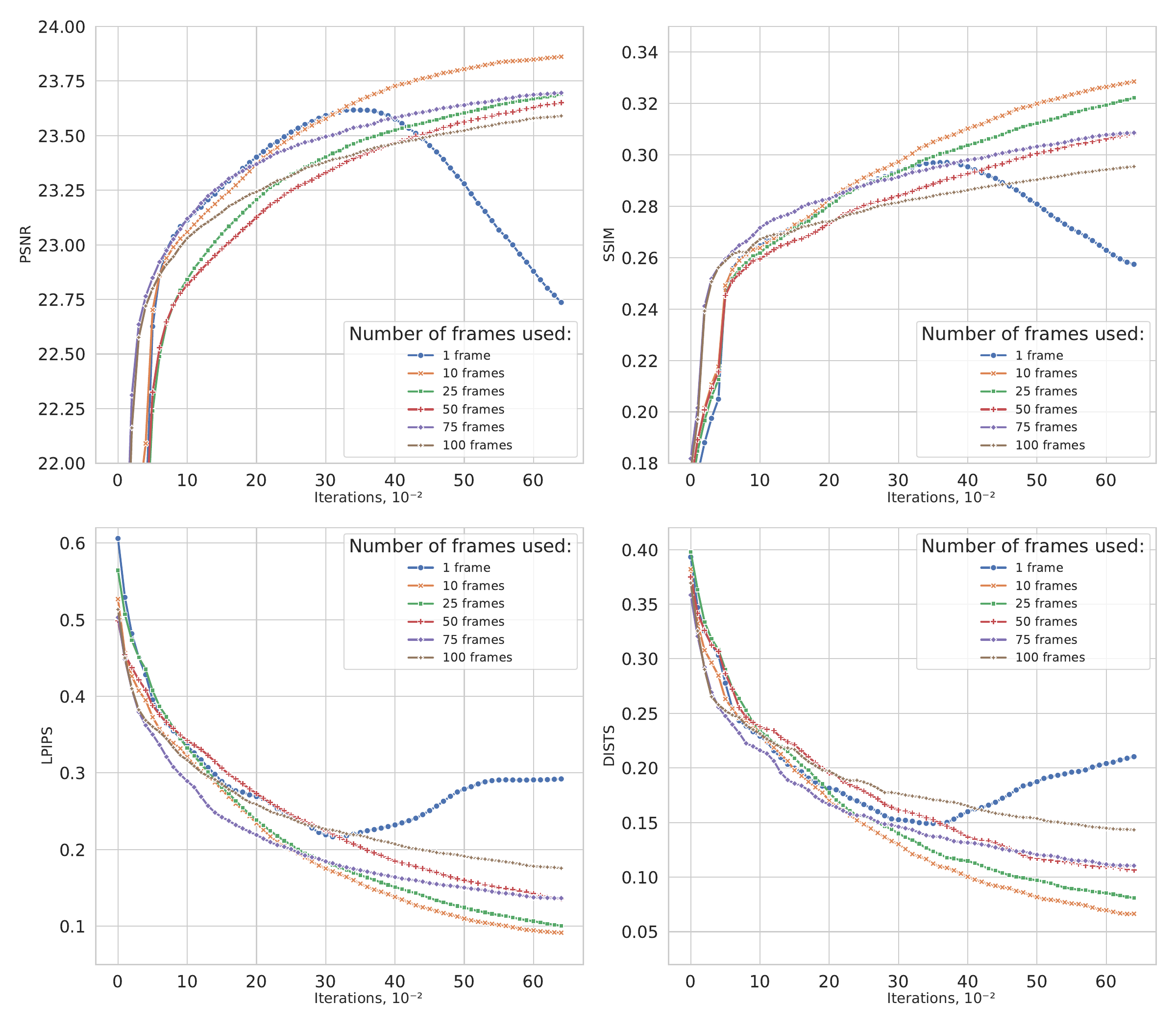}
\caption{\textbf{Comparison of DIPLI reconstruction quality with different number of received LQ frames.} While multiple frames provide complementary information, excessive inputs cause the model to fit inter-frame variations, reducing precision. Optimal performance occurs with approximately 7-13 frames.}
\label{f:n_frames}
\end{figure}

\subsection{Experimental Setup}

The backbone of DIPLI is a four-stage U-Net with $128$ feature channels per stage, instance normalization, and symmetric skip connections between encoder and decoder paths. Each stage consists of two $3 \times 3$ convolutions with ReLU activations and dropout, with downsampling performed via pooling and upsampling via bilinear interpolation followed by feature projection. The output is produced through a multi-step upsampling head that restores spatial resolution by a selected scaling factor $s \in \{ 2, 4, 8 \}$ and applies a final convolution with sigmoid activation.

The evaluation compares DIPLI with the original DIP~\cite{Ulyanov_2020} and modern deep learning models RVRT and DiffIR2VR-Zero, targeting super-resolution by the factor $4$, denoising, and deblurring. DIPLI was trained for $N = 6500$ iterations with a warm-up phase of $n_0 = 6000$, averaging the reconstructions from the last 500 iterations to obtain the final result as a Monte Carlo estimation as described in Algorithm \ref{a:workflow}. The learning rate was set to $\lambda = 0.0025$, with SGLD noise $\sigma_\xi = 0.0025$ and latent perturbation $\sigma_z = 0.01$. The optimal number of source low-quality (LQ) images was determined to be $K = 11$, corresponding to a pivot image preceded by five temporally adjacent frames and followed by five subsequent frames. This value was obtained by testing a range of input sizes from 1 to 100, with results shown in Figure~\ref{f:n_frames}, where the highest reconstruction accuracy is observed for ten neighboring frames. All experiments were conducted on a single NVIDIA V100 GPU; the full pipeline completes in approximately 3-5 minutes per scene.

\subsection{Experiments and Results}

\paragraph{Motion Compensation.}
Numerous approaches exist for motion compensation.
The simplest estimates a single shift vector between two images, offering high speed and sufficient accuracy for scenarios dominated by uniform spatial motion, such as camera shake. In most cases, this single vector adequately models the displacement. However, when the subject moves or heat-induced distortions occur, a global shift becomes insufficient. Optical flow addresses this limitation by extending the single-vector concept to a dense vector field, assigning an individual motion vector to each pixel to capture localized movements.

\begin{figure}[h!]
\centering
\includegraphics[width=0.9\linewidth]{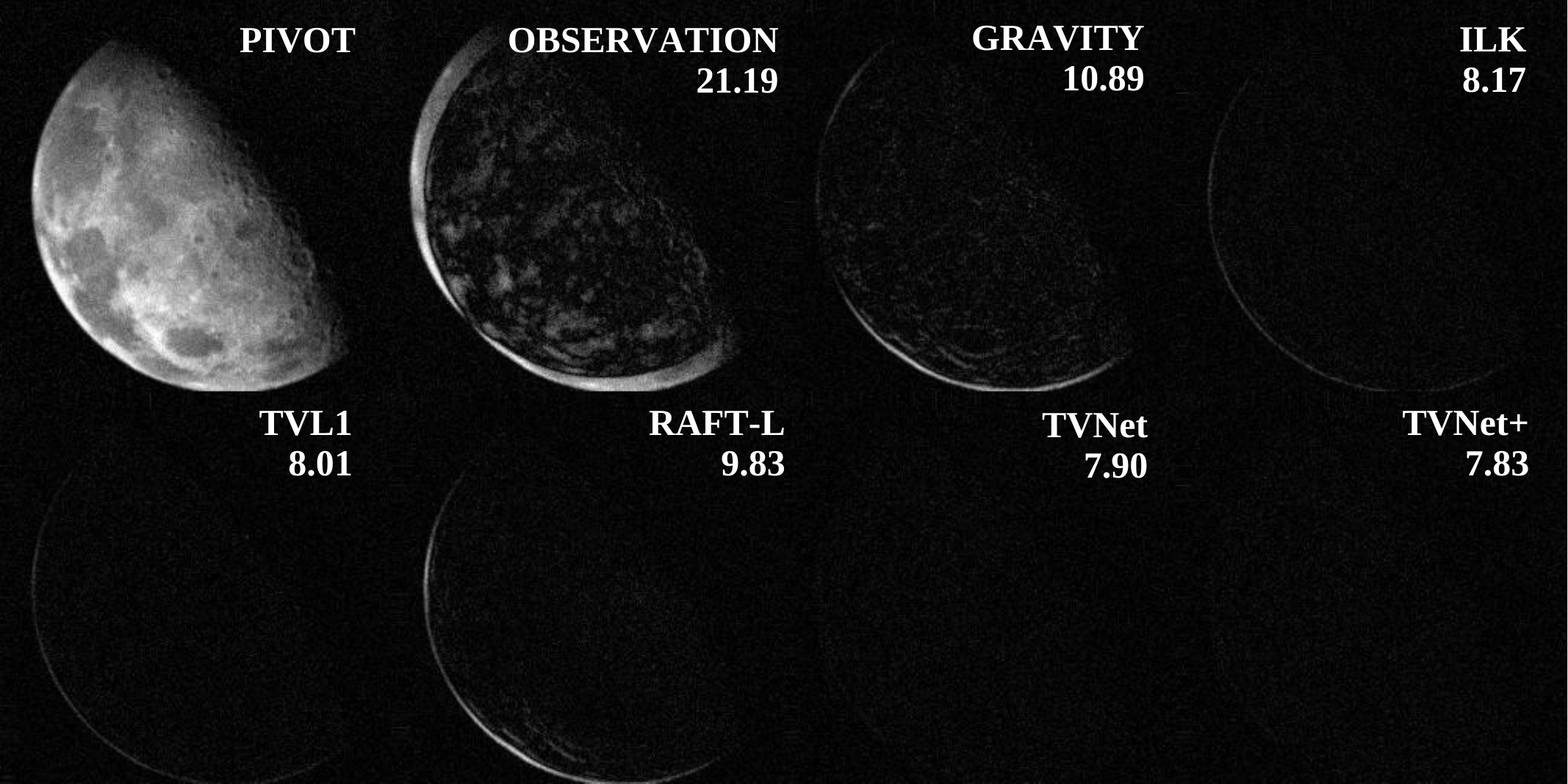}
\caption{\textbf{Comparison of several optical flow computation methods for a pair of images.} The basis image PIVOT is depicted on the top left and followed by several maps of pixel-wise error between the pivot image and the compensated observation obtained with the corresponding methods.}
\label{f:optical_flow}
\end{figure}

There are several possible ways to construct an optical flow with probably the most common way, the TV-L1~\cite{Sanchez2013}. Unfortunately, noisy images significantly hinder optical flow estimation by corrupting the source data and obscuring the true pixel displacement map. To address such challenges, neural network-based approaches, such as TVNet, have been developed and implemented. Without training, TVNet mirrors the behavior of TV-L1, but with additional unsupervised tuning, it can overcome noise. Figure~\ref{f:optical_flow} presents experimental results comparing different motion compensation methods: GRAVITY, directly calculating a spatial shift between two centres of masses in the images, classical Iterative Lucas-Kanade (ILK) method~\cite{ILK1981}, basic TV-L1 method, and two neural network methods RAFT-L~\cite{teed2020} and TVNet. To assess performance, each method estimates a transformation between an image pair (source, target). The transform warps the source to the target, and similarity is quantified by the mean absolute error (MAE) between the warped source and the target; lower MAE indicates better alignment. TVNet achieves the lowest MAE (7.90), and residual registration errors are further mitigated by the confidence-weighted loss (Section~\ref{sec:method}).

\paragraph{SGLD strength.}
UNNP models are prone to overfitting, often memorizing image noise without explicit regularization. To mitigate this, the method adopts Bayesian inference via SGLD MCMC~\cite{Cheng_2019}. SGLD injects Gaussian noise into the parameter updates at each step, promotes posterior exploration, and prevents collapse to a single overfitted solution; averaging late-iteration samples further improves robustness to observation noise.

\begin{figure}[h!]
\centering
\includegraphics[width=0.9\linewidth]{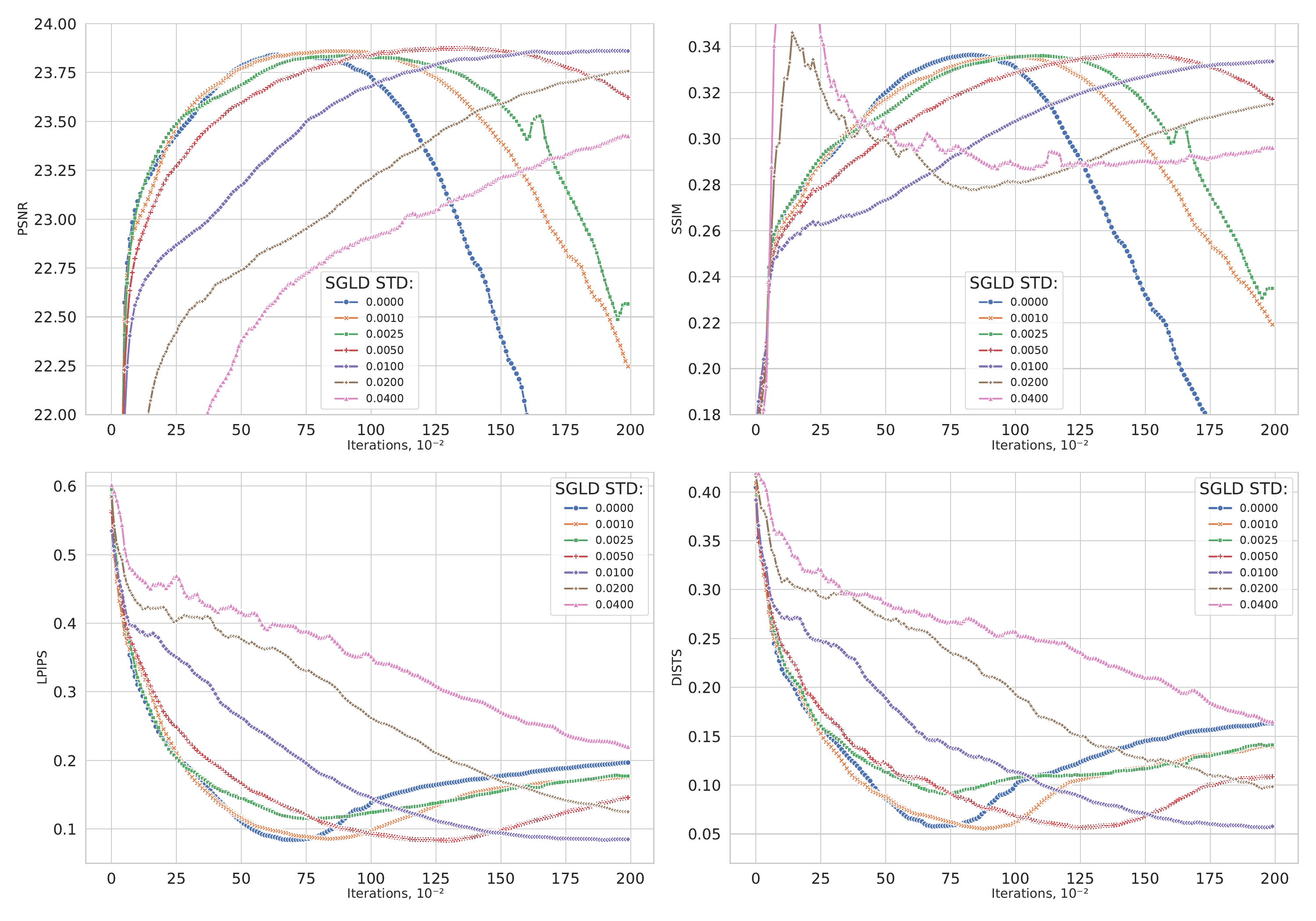}
\caption{\textbf{Comparison of DIPLI reconstruction quality with different values of the SGLD strength coefficient} $\sigma_{\xi}$. Following the approach of~\cite{Cheng_2019}, the results indicate a tradeoff between noise impedance and the number of iterations required for convergence. Higher $\sigma_{\xi}$ values increase the convergence time. For astronomical images in this dataset, the optimal setting was $\sigma_{\xi} = 0.0025$ with 6500 iterations.
}
\label{sgld_plots}
\end{figure}

Regularization strength (i.e., the SGLD noise scale) requires careful tuning: too small has negligible effect, whereas too large impedes learning. Experiments in Figure~\ref{sgld_plots} indicate an optimal noise standard deviation of $\sigma_{\xi}=0.0025$ (equal to the learning rate). Multi-chain convergence diagnostics confirm approximate convergence across representative scenes (Supplementary Note~2).

In addition to SGLD, a small perturbation is applied when sampling the latent vector $z$ from the prior. Although Bayesian DIP reports limited benefit from this perturbation~\cite{Cheng_2019}, the pipeline retains it to promote latent-space smoothness and potential robustness to adversarial latent perturbations, consistent with DIP and randomized smoothing~\cite{Ulyanov2018DeepIP,CohenRandSmoothing}.

\paragraph{Ablation scope.} The main ablation studies focus on $K$ (Figure~\ref{f:n_frames}), $\sigma_\xi$ (Figure~\ref{sgld_plots}), and the registration method (Figure~\ref{f:optical_flow}). Extended ablations of the confidence weighting parameter $\alpha$ across all 12 scenes (Supplementary Note~1) and of the frame count over $K \in \{1,3,5,7,11,15,25,50\}$ across 4 scenes (Supplementary Note~3) are provided in the Supplementary Information.

\begin{figure}[h!]
\centering
\includegraphics[width=0.9\linewidth]{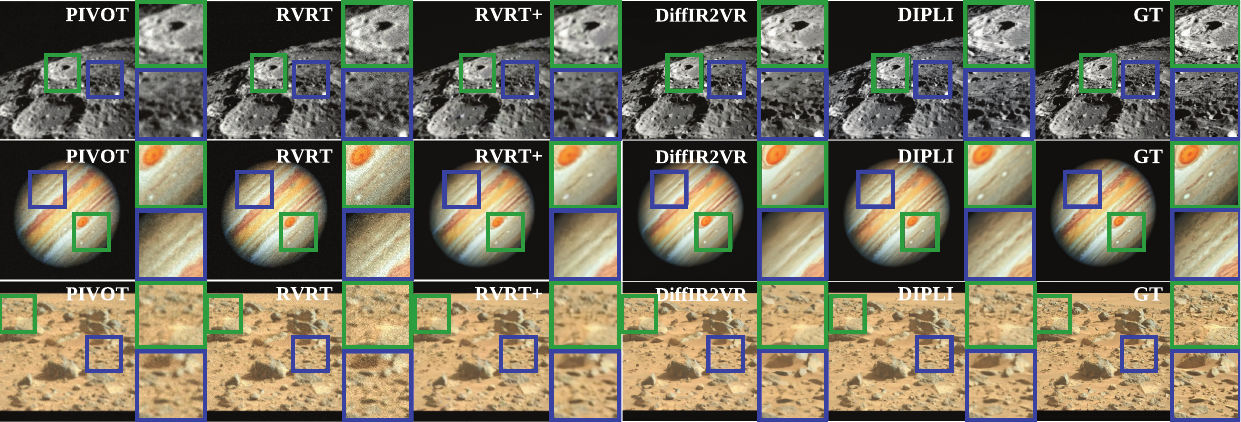}
\caption{\textbf{Synthetic-data reconstructions.} The LQ image pivot is shown on the left, and the remaining images present the corresponding HQ restorations. Insets are included to highlight specific details.}
\label{f:synthetic_data}
\end{figure}

\paragraph{Synthetic Data.}
Figure~\ref{f:synthetic_data} and Table~\ref{t:dl_metrics} indicate that DIPLI achieves the strongest perceptual fidelity (best LPIPS in 12/12 and DISTS in 10/12 scenes), while DiffIR2VR-Zero leads on pixel-level distortion (best PSNR in 9/12 and SSIM in 8/12 scenes). Both DIPLI and DIP are unsupervised and tuned on the target data, whereas RVRT uses pretrained natural-video weights and DiffIR2VR-Zero operates zero-shot by design.

The baseline RVRT configuration exhibits overfitting, visible as fine-grained speckle in the reconstructions. RVRT+ applies preliminary denoising using the same architecture with different weights to reduce direct noise reproduction but yields only modest improvement over the original low-quality (LQ) input. DiffIR2VR-Zero, while producing sharp and visually appealing images, frequently smooths regions and removes salient details from the data. The diffusion-based model also tends to hallucinate textures. Such emphasis on aesthetics over fidelity is problematic in astrophotography, where preserving fine structure is essential to avoid misleading interpretations.

A direct, controlled comparison to DIP is nontrivial because DIP relies on early stopping to prevent overfitting, and the optimal stopping iteration is data-dependent and unknown without ground truth. For the synthetic experiments, a single stopping iteration (470) was selected by scanning iteration counts over the dataset and choosing the value that maximised average PSNR and SSIM; this fixed iteration was then applied uniformly to all DIP runs. Across all metrics, DIPLI consistently outperforms this baseline, isolating the benefit of multi-frame fusion, confidence weighting, and SGLD sampling. On real data, the absence of references makes such tuning infeasible, leaving the DIP stopping point indeterminate.

\begin{table*}[h!]
\centering
\begin{tabular}{|l|c|c|c|c|c|c|c|c|c|c|}
\hline
        & \multicolumn{5}{c|}{PSNR} & \multicolumn{5}{c|}{SSIM} \\
\hline
        Dataset & DIP & RVRT & RVRT+ & DiffIR2VR & DIPLI & DIP & RVRT & RVRT+ & DiffIR2VR & DIPLI \\
\hline
\verb|01| & 20.57 & 19.90 & \textit{21.20} & 20.79 & \textbf{22.25} & 0.39 & 0.31 & 0.45 & \textbf{0.54} & \textbf{0.54} \\
\verb|02| & 28.35 & 23.64 & 29.95 & \textbf{32.63} & \textit{31.19} & 0.42 & 0.27 & 0.50 & \textbf{0.67} & \textit{0.52} \\
\verb|03| & 23.75 & 21.77 & \textbf{24.33} & \textit{24.24} & 23.93 & 0.30 & 0.27 & 0.31 & \textbf{0.52} & \textit{0.34} \\
\verb|04| & 28.94 & 23.68 & 30.26 & \textbf{31.13} & \textit{30.84} & 0.61 & 0.29 & 0.72 & \textbf{0.77} & \textit{0.75} \\
\verb|05| & 27.16 & 23.15 & 28.53 & \textbf{31.39} & \textit{28.94} & 0.58 & 0.29 & 0.71 & \textbf{0.77} & \textit{0.74} \\
\verb|06| & 26.50 & 23.07 & \textit{27.82} & \textbf{30.38} & 26.79 & 0.36 & 0.26 & 0.42 & \textbf{0.64} & \textit{0.43} \\
\verb|07| & 26.05 & 22.97 & \textit{27.25} & \textbf{28.30} & 26.19 & 0.33 & 0.27 & 0.36 & \textbf{0.56} & \textit{0.38} \\
\verb|08| & 21.81 & 20.59 & 22.27 & \textbf{22.52} & \textit{22.51} & 0.40 & 0.29 & 0.45 & \textit{0.48} & \textbf{0.52} \\
\verb|09| & 24.59 & 22.21 & \textit{25.41} & \textbf{26.32} & 24.86 & 0.29 & 0.27 & 0.30 & \textbf{0.52} & \textit{0.33} \\
\verb|10| & 25.78 & 22.85 & \textit{26.79} & \textbf{27.60} & 26.18 & 0.54 & 0.31 & 0.64 & \textit{0.64} & \textbf{0.67} \\
\verb|11| & 26.06 & 22.88 & \textit{27.10} & \textbf{29.24} & 25.89 & 0.59 & 0.32 & 0.71 & \textbf{0.74} & \textit{0.73} \\
\verb|12| & 26.28 & 22.76 & \textit{27.22} & \textbf{28.62} & 26.16 & 0.53 & 0.28 & 0.62 & \textbf{0.65} & \textit{0.64} \\
\hline
\end{tabular}
\label{t:classic_metrics}

\vspace{2mm}

\centering
\begin{tabular}{|l|c|c|c|c|c|c|c|c|c|c|}
\hline
        & \multicolumn{5}{c|}{DISTS} & \multicolumn{5}{c|}{LPIPS} \\
\hline
        Dataset & DIP & RVRT & RVRT+ & DiffIR2VR & DIPLI & DIP & RVRT & RVRT+ & DiffIR2VR & DIPLI \\
\hline
\verb|01| & 0.17 & 0.21 & 0.16 & \textit{0.12} & \textbf{0.05} & 0.32 & 0.34 & 0.21 & \textit{0.20} & \textbf{0.08} \\
\verb|02| & 0.28 & 0.30 & 0.18 & \textit{0.16} & \textbf{0.08} & 0.39 & 0.43 & \textit{0.15} & 0.19 & \textbf{0.05} \\
\verb|03| & 0.21 & 0.24 & 0.15 & \textit{0.15} & \textbf{0.07} & 0.32 & 0.35 & \textit{0.21} & 0.23 & \textbf{0.10} \\
\verb|04| & 0.22 & 0.28 & \textit{0.13} & 0.14 & \textbf{0.09} & 0.28 & 0.34 & 0.12 & \textit{0.12} & \textbf{0.08} \\
\verb|05| & 0.22 & 0.27 & 0.16 & \textit{0.12} & \textbf{0.11} & 0.35 & 0.40 & 0.18 & \textit{0.16} & \textbf{0.11} \\
\verb|06| & 0.27 & 0.30 & 0.17 & \textit{0.13} & \textbf{0.08} & 0.42 & 0.49 & 0.18 & \textit{0.15} & \textbf{0.09} \\
\verb|07| & 0.24 & 0.27 & 0.14 & \textit{0.10} & \textbf{0.07} & 0.37 & 0.43 & 0.18 & \textit{0.16} & \textbf{0.08} \\
\verb|08| & 0.27 & 0.30 & 0.21 & \textit{0.13} & \textbf{0.12} & 0.36 & 0.42 & 0.29 & \textit{0.21} & \textbf{0.14} \\
\verb|09| & 0.23 & 0.25 & 0.13 & \textit{0.12} & \textbf{0.08} & 0.34 & 0.36 & \textit{0.17} & 0.19 & \textbf{0.09} \\
\verb|10| & 0.28 & 0.32 & 0.19 & \textbf{0.14} & \textit{0.15} & 0.39 & 0.48 & 0.25 & \textit{0.20} & \textbf{0.15} \\
\verb|11| & 0.27 & 0.30 & 0.18 & \textbf{0.10} & \textit{0.14} & 0.40 & 0.48 & 0.21 & \textit{0.14} & \textbf{0.13} \\
\verb|12| & 0.23 & 0.29 & 0.18 & \textit{0.17} & \textbf{0.15} & 0.34 & 0.46 & 0.27 & \textit{0.23} & \textbf{0.16} \\
\hline
\end{tabular}
\caption{\textbf{Comparison of reconstruction metrics for synthetic data.}
PSNR, SSIM, DISTS, and LPIPS are reported for the observation and ground truth, alongside results from DIPLI, RVRT, and DiffIR2VR-Zero. The DIP column corresponds to the original Deep Image Prior~\cite{Ulyanov_2020} with an early stopping point fixed for the entire dataset (470 iterations). \textbf{Best} and \textit{second-best} results are highlighted in bold.}
\label{t:dl_metrics}
\end{table*}

\paragraph{Real Data}

\begin{figure}[t!]
\centering
\includegraphics[width=1\linewidth]{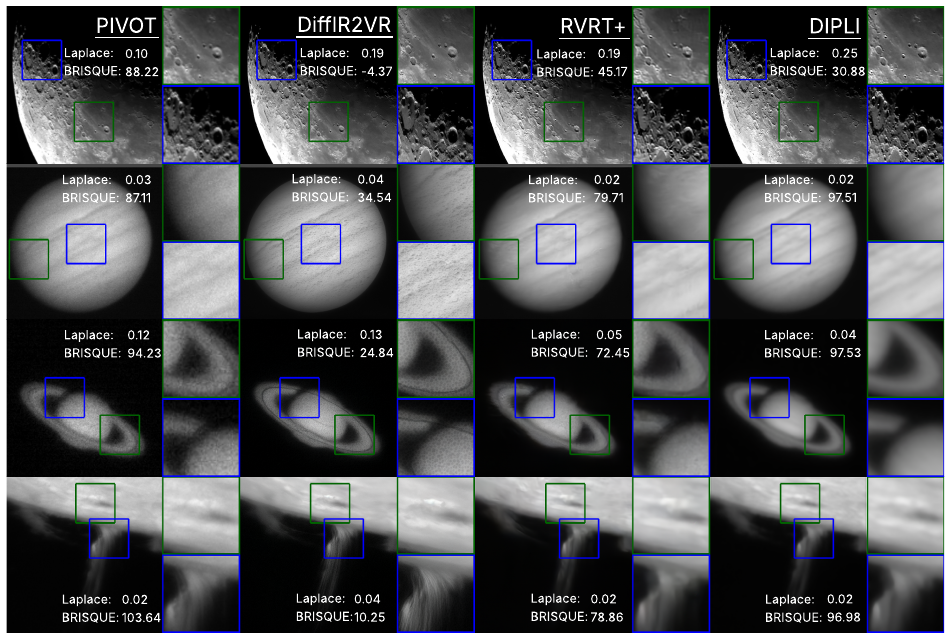}
\caption{\textbf{Real-data reconstructions for different celestial objects.} The LQ image pivot is shown on the left, and the remaining images present the corresponding HQ restorations. Insets are included to highlight specific details. Laplace Energy and BRISQUE metrics are reported. Higher values are better for Laplace Energy, and lower values are better for BRISQUE.}
\label{f:real_data}
\end{figure}

In the absence of ground truth, full-reference metrics such as PSNR, SSIM, LPIPS, and DISTS, are inapplicable. Figure~\ref{f:real_data} therefore presents qualitative comparisons together with two no-reference indicators: Laplacian energy and BRISQUE~\cite{mittal2012no}. Laplacian energy measures high-frequency content and is widely used in astronomical imaging as a sharpness or focus measure; it provides a reasonable sanity check but is noise-sensitive and cannot by itself distinguish genuine detail from hallucinated structure. BRISQUE, a Natural Scene Statistics (NSS)-based model trained on natural images, exhibits poor transferability to astronomical scenes and produced unstable rankings under distribution shift and compound distortions. Accordingly, the analysis prioritizes visual assessment (with limited expert inspection as a sanity check), treats Laplacian energy as auxiliary evidence, and de-prioritizes BRISQUE for this domain.

Qualitative assessment indicates that DIPLI produces clean reconstructions with strong detail retention, effective noise suppression, and no visually apparent method-induced artifacts. By contrast, RVRT+ frequently exhibits fine-grained speckle and residual ringing, while DiffIR2VR-Zero yields perceptually sharp results that nonetheless over-smooth regions and occasionally hallucinate texture near high-frequency structures. These observations align with the synthetic-data benchmarks with ground truth (Figure~\ref{f:synthetic_data}, Table~\ref{t:dl_metrics}).

We note that the current real-world evaluation focuses on resolved solar-system objects with sufficient spatial texture for optical flow estimation. Point-source fields and extended low-surface-brightness objects represent known boundary conditions; the Supplementary Information probes these through a failure case analysis on a synthetic star field (Supplementary Note~4) and reconstructions of six Zwicky Transient Facility~\cite{Bellm2019ZTF} targets (Supplementary Note~5).

%% file: chapters/conclusion.tex
\section{Conclusion}
\label{sec:conclusion}

This work introduced \textbf{DIPLI}, an enhanced DIP-based framework designed for astrophotography applications. It addresses the unique challenges posed by limited training data by aggregating multiple frames via back-projection, aligning them using TVNet-based optical flow, and regularizing optimization through a Bayesian scheme based on stochastic gradient Langevin dynamics with Monte Carlo averaging. 

Experiments on synthetic data with ground truth demonstrate that DIPLI achieves competetive perceptual fidelity scores relative to DIP, RVRT, and DiffIR2VR-Zero: best LPIPS in 12 out of 12 and best DISTS in 10 out of 12 tested scenes. In traditional distortion metrics (PSNR, SSIM), diffusion-based methods achieve higher values, aligning with the well-known perceptual-distortion trade-off in image restoration~\cite{Blau2018}. Ablation studies validate the three core design decisions: multi-frame fusion ($K$ ablation), Bayesian regularization ($\sigma_\xi$ ablation), and dense optical flow registration (method comparison), each showing clear and interpretable effects on reconstruction quality. In real scenes without ground-truth references, qualitative visual inspection suggests that DIPLI produces reconstructions with apparent detail preservation and reduced artifacts, consistent with the synthetic-data findings; however, these observations are inherently limited by the absence of reference-based evaluation and should be interpreted as an illustrative proof-of-concept rather than rigorous validation.

\paragraph{Limitations.}
Several limitations should be noted when considering the applicability of DIPLI. First, the current evaluation is restricted to $256 \times 256$ input resolution; scalability to wide-field imaging or large mosaics has not been tested and may require modifications to the mini-batch strategy and memory management. Second, the method depends on the GPU hardware, which may limit accessibility for some users. Third, the optical flow estimation via TVNet assumes sufficient spatial texture in the input frames; scenarios involving point sources (star fields) or very low surface brightness objects may not provide adequate features for reliable flow estimation and require a different approach for motion compensation. Finally, the real-world evaluation is limited to resolved solar-system objects, and generalization to other astronomical domains remains to be validated.

\paragraph{Future work.}
Future directions include exploring more efficient network architectures to reduce computational requirements, investigating adaptive strategies for frame selection under varying turbulence conditions, and extending the framework to handle point-source fields through alternative registration approaches. Expanding the scope of validation to diverse astronomical targets and larger field sizes would further establish the practical utility of the method.